\documentclass[10pt]{report}
\usepackage[authoryear,sort&compress,round]{natbib}

\usepackage[utf8]{inputenc} %
\usepackage[T1]{fontenc}    %

\usepackage{parskip}        %
\usepackage{url}            %
\usepackage{booktabs}       %
\usepackage{amsfonts}       %
\usepackage{nicefrac}       %
\usepackage{microtype}      %
\usepackage{xcolor}         %
\usepackage[dvipsnames]{xcolor} %
\usepackage{graphicx}
\usepackage{animate}        %
\usepackage{subcaption}
\usepackage{tabularx}
\usepackage{makecell}
\usepackage{adjustbox}
\usepackage{setspace}
\usepackage{todonotes}
\usepackage{colortbl}
\usepackage{wrapfig}

\newcolumntype{M}[1]{>{\centering\arraybackslash}m{#1}}
\usepackage{float}
\usepackage{tikz}
\usetikzlibrary{positioning,shapes,arrows}
\usepackage{amsmath,amsfonts,bm, bbm,leftindex}
\usepackage{multirow}
\usepackage{comment}
\usepackage{gensymb}
\usepackage{lipsum}
\usetikzlibrary{arrows.meta, positioning, fit}
\usepackage[para]{threeparttable}
\usepackage{tikz}
\usetikzlibrary{tikzmark}

\def\eqref#1{equation~\ref{#1}}

\def\1{\bm{1}}

\DeclareMathAlphabet{\mathsfit}{\encodingdefault}{\sfdefault}{m}{sl}
\SetMathAlphabet{\mathsfit}{bold}{\encodingdefault}{\sfdefault}{bx}{n}

\makeatletter
\let\save@mathaccent\mathaccent
\newcommand*\if@single[3]{%
  \setbox0\hbox{${\mathaccent"0362{#1}}^H$}%
  \setbox2\hbox{${\mathaccent"0362{\kern0pt#1}}^H$}%
  \ifdim\ht0=\ht2 #3\else #2\fi
  }
\newcommand*\rel@kern[1]{\kern#1\dimexpr\macc@kerna}
\newcommand*\widebar[1]{\@ifnextchar^{{\wide@bar{#1}{0}}}{\wide@bar{#1}{1}}}
\newcommand*\wide@bar[2]{\if@single{#1}{\wide@bar@{#1}{#2}{1}}{\wide@bar@{#1}{#2}{2}}}
\newcommand*\wide@bar@[3]{%
  \begingroup
  \def\mathaccent##1##2{%
    \let\mathaccent\save@mathaccent
    \if#32 \let\macc@nucleus\first@char \fi
    \setbox\z@\hbox{$\macc@style{\macc@nucleus}_{}$}%
    \setbox\tw@\hbox{$\macc@style{\macc@nucleus}{}_{}$}%
    \dimen@\wd\tw@
    \advance\dimen@-\wd\z@
    \divide\dimen@ 3
    \@tempdima\wd\tw@
    \advance\@tempdima-\scriptspace
    \divide\@tempdima 10
    \advance\dimen@-\@tempdima
    \ifdim\dimen@>\z@ \dimen@0pt\fi
    \rel@kern{0.6}\kern-\dimen@
    \if#31
      \overline{\rel@kern{-0.6}\kern\dimen@\macc@nucleus\rel@kern{0.4}\kern\dimen@}%
      \advance\dimen@0.4\dimexpr\macc@kerna
      \let\final@kern#2%
      \ifdim\dimen@<\z@ \let\final@kern1\fi
      \if\final@kern1 \kern-\dimen@\fi
    \else
      \overline{\rel@kern{-0.6}\kern\dimen@#1}%
    \fi
  }%
  \macc@depth\@ne
  \let\math@bgroup\@empty \let\math@egroup\macc@set@skewchar
  \mathsurround\z@ \frozen@everymath{\mathgroup\macc@group\relax}%
  \macc@set@skewchar\relax
  \let\mathaccentV\macc@nested@a
  \if#31
    \macc@nested@a\relax111{#1}%
  \else
    \def\gobble@till@marker##1\endmarker{}%
    \futurelet\first@char\gobble@till@marker#1\endmarker
    \ifcat\noexpand\first@char A\else
      \def\first@char{}%
    \fi
    \macc@nested@a\relax111{\first@char}%
  \fi
  \endgroup
}
\makeatother

\definecolor{darkred}{rgb}{0.7, 0.0, 0.0}

\usepackage{amsmath} 
\usepackage{authblk}

\usepackage[table]{xcolor}

\usepackage{pifont}
\usepackage{graphicx}

\usepackage[nameinlink]{cleveref}
\crefname{equation}{Eq.}{Eqs.}
\crefname{figure}{Fig.}{Figs.}
\crefname{section}{Sec.}{Sec.}
\crefname{appendix}{App.}{App.}
\crefname{table}{Tab.}{Tabs.}
\crefname{algorithm}{Algo}{Algo}
\crefname{thm}{Thm}{Thm}
\Crefname{thm}{Thm}{Thm}
\crefname{prop}{Prop}{Prop}

\newcommand{\crefnames}[3]{%
  \@for\next:=#1\do{%
    \expandafter\crefname\expandafter{\next}{#2}{#3}%
  }%
}

\title{StableIDM: Stabilizing Inverse Dynamics Model against Manipulator Truncation via Spatio-Temporal Refinement}

\renewcommand*{\Affilfont}{\small\centering}

\author[1,2]{Kerui Li}
\author[3]{Zhe Jing}
\author[1]{Xiaofeng Wang}
\author[1]{Zheng Zhu}
\author[1]{Yukun Zhou}
\author[1]{Guan Huang}
\author[2]{Dongze Li}
\author[3]{Qingkai~Yang}
\author[2]{Huaibo Huang}

\affil[1]{GigaAI}
\affil[2]{Institute of Automation, Chinese Academy of Sciences}
\affil[3]{Beijing Institute of Technology}

\begin{document}
\maketitle

\begin{center}
    \centering
    \captionsetup{type=figure, justification=justified, singlelinecheck=false}
    \includegraphics[width=0.95\linewidth]{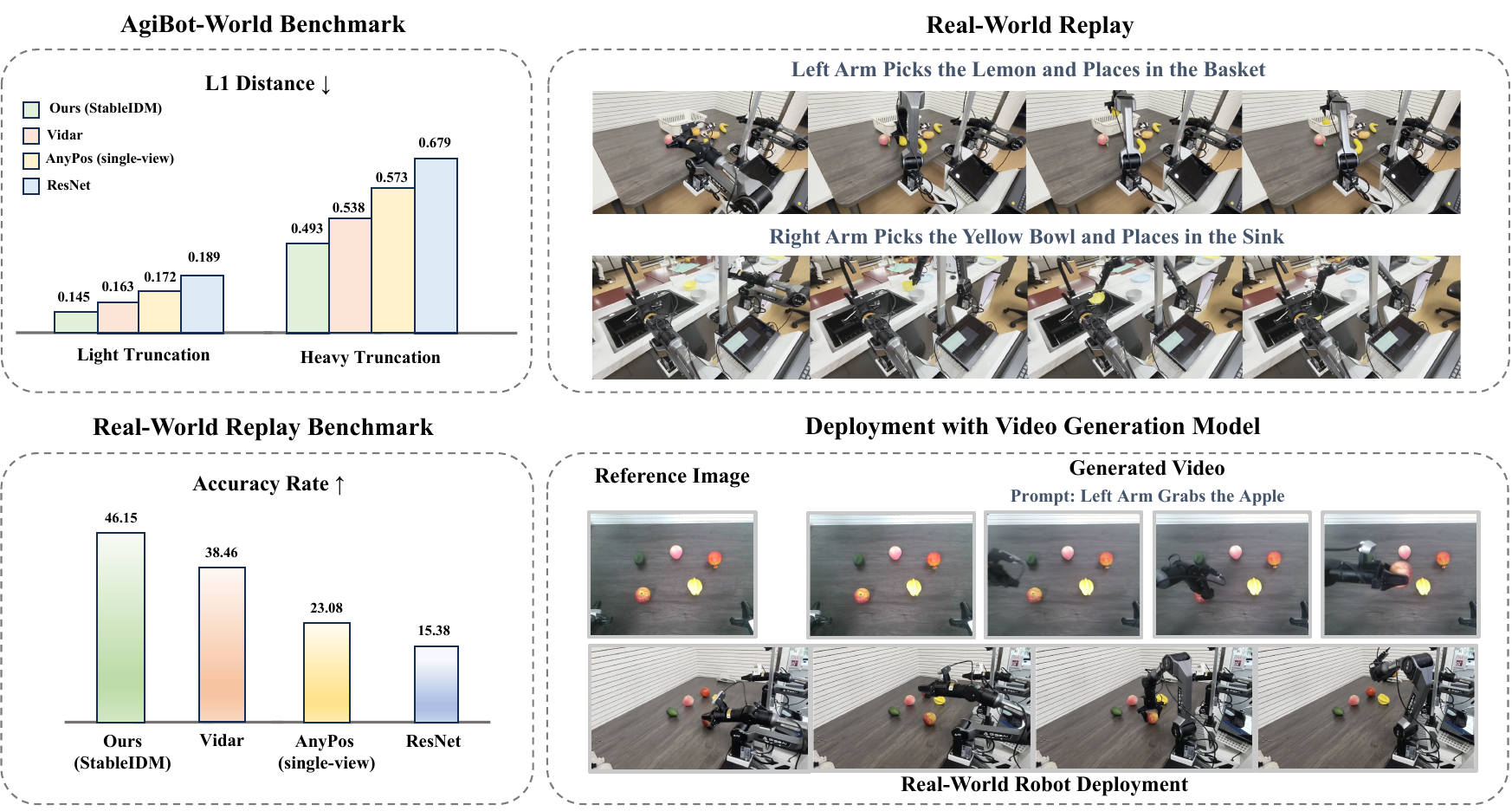}
\vspace{-0.1in}
    \caption{The proposed \textit{StableIDM} addresses manipulator truncation in inverse dynamics models using a spatio-temporal feature refinement framework. On the AgiBot World benchmark (top left), it attains the lowest L1 distance under both light and heavy truncation, with especially large gains in the heavy regime where baselines degrade sharply. This offline robustness carries over to real world experiments: \textit{StableIDM} achieves higher replay success rates (bottom left), produces smoother trajectories (top right), and reliably decodes video generated plans into executable robot actions (bottom right).}
    \label{fig:teaser}
\end{center}

\begin{abstract}
\vspace{-0.1in}
Inverse Dynamics Models (IDMs) map visual observations to low-level action commands, serving as central components for data labeling and policy execution in embodied AI. However, their performance degrades severely under manipulator truncation, a common failure mode that makes state recovery ill-posed and leads to unstable control. We present \textit{StableIDM}, a spatio-temporal framework that refines features from visual inputs to stabilize action predictions under such partial observability. \textit{StableIDM} integrates three complementary components: (1) auxiliary robot-centric masking to suppress background clutter, (2) Directional Feature Aggregation (DFA) for geometry-aware spatial reasoning, which extracts anisotropic features along directions inferred from the visible arm and (3) Temporal Dynamics Refinement (TDR) to smooth and correct predictions via motion continuity. Extensive evaluations validate our approach: \textit{StableIDM} improves strict action accuracy by 12.1\% under severe truncation on the AgiBot benchmark, and increases average task success by 9.7\% in real-robot replay. Moreover, it boosts end-to-end grasp success by 11.5\% when decoding video-generated plans, and improves downstream VLA real-robot success by 17.6\% when functioning as an automatic annotator. These results demonstrate that \textit{StableIDM} provides a robust and scalable backbone for both policy execution and data generation in embodied artificial intelligence.
\end{abstract}

\section{Introduction}

Recent progress in Vision–Language–Action (VLA) models~\cite{gigabrain0,walloss,g0,pi0,pi0.5,gr3} has rapidly advanced embodied intelligence by scaling policies with Internet-scale vision–language corpora and large cross-robot datasets. 
Bridging the gap between high-level perception and low-level robot control is a core challenge in embodied intelligence~\cite{bu2024closed,unipi,unleashing,hoque2025egodex,yuan2025roboengine}. 
Inverse Dynamics Models (IDMs) map visual observations to executable action commands, providing a practical mechanism for bridging this perception–control gap.
This capability makes IDMs key components in modern robotics pipelines. 
As automatic action annotators, IDMs~\cite{dreamgen,wow} annotate large-scale videos for training VLA models.
As policies, IDMs~\cite{unipi,susie,robodreamer,anypos,vidar} translate visual plans from video generation models into real robot control commands.

Despite their widespread utility, IDMs are highly sensitive to manipulator truncation, where parts of the robot arm move out of the camera view and only a short segment remains visible.
On the AgiBot World benchmark, \cref{fig:teaser} shows that the L1 distance between predicted and ground truth actions is much larger under heavy truncation than under light truncation for all methods, while accuracy drops at the same time.
This pattern indicates that once most of the manipulator disappears from the image, IDMs struggle to recover the underlying state and their action predictions become unreliable. The core difficulty is that truncation removes key structural and geometric cues, so the model can no longer infer the full manipulator configuration from a single frame. As a result, the inverse dynamics regression becomes highly ill posed and small visual changes can lead to large errors and unstable control.

To restore stability under manipulator truncation, we propose \textit{StableIDM}, an IDM built on spatio-temporal feature refinement. The system is a three-stage pipeline: 
(i) \textbf{Robot-centric Masking} suppresses background clutter so the encoder attends to the robot manipulator.
(ii) \textbf{Directional Feature Aggregation (DFA)} consolidates orientation-sensitive cues into a robust, direction-aware descriptor that preserves geometry under partial observability.
(iii) \textbf{Temporal Dynamics Refinement (TDR)} enforces causal smoothing and completes short-term dynamics via a two-step temporal stack: temporal fusion before DFA repairs the current visual features using adjacent frames, and a lightweight temporal regressor after DFA stabilizes the final action.
During inference, \textit{StableIDM} runs on a fixed-length sliding window and outputs the current action.

We validate \textit{StableIDM} with four experiments including offline action prediction, real-world replay, video-plan deployment with video generation and data generation for VLA training.
All metrics indicate that \textit{StableIDM} offers strong performance for both large-scale action labeling and policy execution.
First, on the AgiBot benchmark, \textit{StableIDM} improves offline action prediction accuracy by 12.1\% on its severe truncation subset compared to prior methods. Second, in real-robot experiments, it increases the average task success rate by 9.7\%. Third, when integrated into a video generation-IDM decoding-robot execution policy pipeline, total grasp success rates increase by 11.5\%. Finally, using \textit{StableIDM} to automatically annotate VLA training data improves the success rate of downstream real-robot tasks by 17.6\% compared with VLA trained on pure real-world collected data.

Our main contributions are as follows: \begin{itemize} 
\item We propose \textit{StableIDM}, a spatio-temporal IDM framework, to achieve stable action predictions under manipulator truncation. 
\item Our core design is an IDM architecture including three modules: 
a) Robot-centric Masking for noise suppression, b) Directional Feature Aggregation (DFA) for geometric reasoning and c) Temporal Dynamics Refinement (TDR) for motion continuity.
\item Comprehensive validation demonstrating strong performance in offline prediction, real-world replay, online policy deployment, and as a high-fidelity data annotator for VLA models.
\end{itemize}
\section{Related Works}

IDM performs action recovery from videos. This architecture admits two applications: (i) an offline use where it acts as a scalable action annotator that converts internet or generated videos into (video, action) pairs for training large VLA models (\cref{sec:rw1}), and (ii) an online use where IDM is combined with a video generation model as a deployable policy. The policy executes tasks by generating videos, then extracting and replaying actions from them (\cref{sec:rw2}).

\subsection{IDMs as Action Annotators}
\label{sec:rw1}

A core bottleneck in embodied AI is the scarcity of scalable, diverse, and semantically aligned action data. Existing sources include three main types. (i) Simulation data scales (e.g., RoboTwin~\cite{robotwin1}, ManiBox~\cite{manibox} and so on~\cite{robotwin2,agibot,robomind}) but suffers from Sim2Real gaps and limited physical fidelity. 
(ii) Real-robot data (e.g., Diffusion Policy~\cite{diffusionpolicy,diffusionpolicy2,unisim}, Mobile Aloha~\cite{aloha1,aloha2}, and other manipulation datasets~\cite{rdt,bridge,openxembodiment,agibot,ACT}) are practical but expensive and task-specific. 
(iii) Internet videos offer rich priors with promising early results~\cite{gr2,robodreamer} yet lack precise action labels needed for direct policy training. This scarcity constrains action-centric policies, especially end-to-end VLA models (e.g. RT-X~\cite{openxembodiment}, Octo~\cite{octo}, OpenVLA~\cite{openvla}).

To expand supervision beyond costly real-world trajectories, we treat the IDM as a scalable action annotator that converts video-only data into training pairs at scale. Concretely, large volumes of trajectories are obtained either by collecting internet videos or by generating synthetic videos, and either a learned world model or a decoupled IDM recovers actions and produces pseudo-labels. The resulting (video, action) pairs are used to train downstream visuomotor or VLA policies. Representative label-expansion pipelines include DreamGen~\cite{dreamgen}, which synthesizes labeled robot trajectories, and web-video-conditioned methods such as Gen2Act~\cite{gen2act}, while WoW~\cite{wow} closes an imagination-to-action loop by training a generative world model together with an inverse-dynamics module. Earlier decoupled systems such as UniPi~\cite{unipi}, UniSim~\cite{unisim}, and SuSIE~\cite{susie} provide additional architectural background for separating visual plan generation from action recovery, which we build upon in our labeling-style use of IDMs. Overall, IDM-as-annotator directly addresses the data bottleneck by transforming both internet and generated videos into large-scale, higher-quality supervision for training VLA models.

\subsection{IDMs as Lightweight Policies}
\label{sec:rw2}

Beyond action labeling, IDM can be combined with video generation model to form a deployable policy-like execution pipeline without dense action labels. Given a text prompt and current observations, video generation models generate an execution video plan, and the IDM performs frame inverse dynamics to extract and replay robot actions.

Contemporary video generators can be coarsely grouped into two families that supply these plans: (i) internet-scale text or image conditioned diffusion or latent-diffusion video models~\cite{cogvideo,cogvideox,sora,latentVideoDiffusion,wan,cosmos1,cosmos2,svd,Vidu,hunyuanvideo,genie}.

This idea is instantiated by works that combine video priors with IDM-based execution, such as AnyPos and Vidar~\cite{anypos,vidar}, and by imagination-to-action systems like RoboDreamer~\cite{robodreamer}. While RoboDreamer can in principle also support offline labeling as noted in \cref{sec:rw1}, here we focus on its role as an imagination-to-action system within the online IDM combined with video generation pipeline. The same early decoupled efforts also inform modular execution beyond single-arm or single-view settings. For contrast, tightly coupled video–action approaches (e.g., VPP~\cite{vpp}, UVA~\cite{uva}, VidMan~\cite{vidman}) train action predictors directly on top of video diffusion backbones and, in the case of UVA and VidMan, learn unified latent spaces via end-to-end training, reducing modularity and adaptability relative to the decoupled video generation models and IDMs. In practice, IDM thus plays a dual role: it (i) labels large-scale videos for training VLA models, and (ii) executes generated plans from video generation models as a deployable policy.
\section{Method}
\label{sec:method}

\begin{figure}[t]
    \centering
    \captionsetup{type=figure, justification=justified, singlelinecheck=false}
    \includegraphics[width=0.95\linewidth]{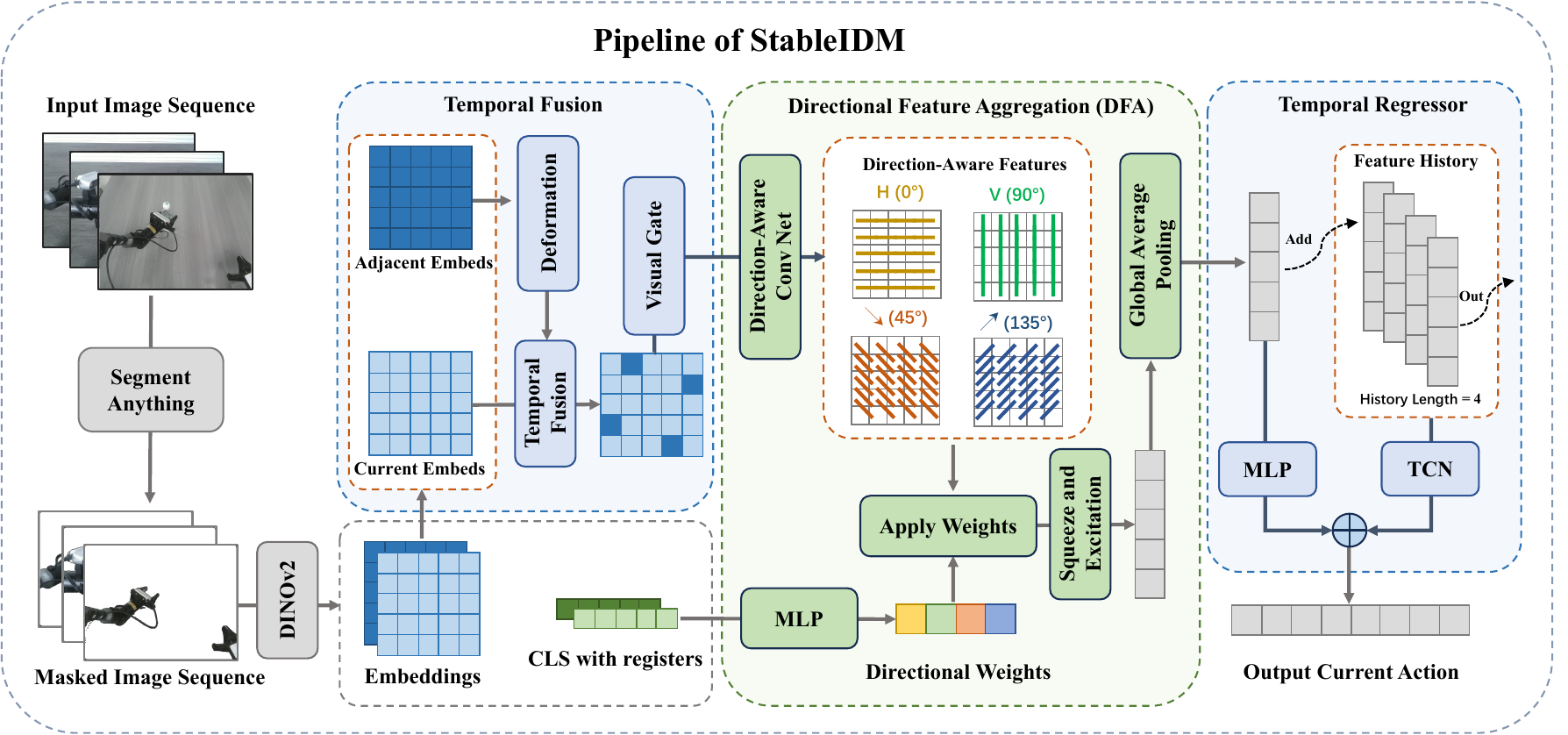}
  \caption{The architecture of \textit{StableIDM}. Our framework is a spatio-temporal IDM designed to enhance feature robustness and refine predictions to ensure stability against manipulator truncation.
(i) Robot-centric Masking: Segment Anything provides robot-centric masking to suppress background noise. The DINO encoder then extracts visual Embeddings and CLS tokens from the masked input sequence.
(ii) DFA: This core spatial component consolidates direction-specific cues. It uses a direction-aware convolution network and adaptive directional weights (derived from CLS tokens via an MLP) to produce a robust and direction-aware feature descriptor via global average pooling.
(iii) TDR: This module operates at both visual and feature levels to refine the representation.
The temporal fusion block enhances the current visual features by integrating information from adjacent embeds and current embeds using a visual gate.
The temporal regressor block smooths the final prediction by analyzing the feature history with a TCN, resulting in a stable output action.} 
  \label{fig:architecture}
\end{figure}

\subsection{Preliminary}
\label{sec:preliminary}

Inverse Dynamics Models (IDMs) are traditionally formulated to recover actions from visual observations. The standard and most common approach, which we refer to as the standard IDM paradigm, formulates this as a single-frame regression $a_t = f_\theta(o_t)$~\cite{vidar,anypos}. However, under manipulator truncation, this standard paradigm suffers from two fundamental weaknesses: First, \textbf{Temporal Memorylessness:} the single-frame model lacks historical context, making the geometric recovery of truncated links highly ill-posed. Second, \textbf{Isotropic Spatial Aggregation:} global self-attention mixes spatial information uniformly, failing to preserve fine-grained, direction-sensitive cues critical for articulated arms. 

To overcome these limitations and achieve stability, we formally define our \textit{StableIDM} as a history-conditioned action decoder $f_\theta$ that maps a brief causal history of visual observations $o_{t-K+1:t}$ to the current action $a_t$. In our setup, the action space $a_t$ is parameterized as a continuous vector comprising the dual-arm joint angles and gripper opening dimensions. Following recent visual planning pipelines (e.g., AnyPos~\cite{anypos}), this causal decoding paradigm explicitly bypasses the classical future-conditioned formulation ($o_t, o_{t+1} \rightarrow a_t$). Rather than learning a closed-loop imitation policy directly, our model is designed to recover low-level actions from streaming camera inputs and decode generated visual plans for real-time open-loop deployment. This architecture is typically implemented with a powerful visual encoder $E_\phi$ (e.g., a ViT~\cite{vit,transformer}) followed by a regression head $h_\psi$ (e.g., a Multi-Layer Perceptron (MLP)~\cite{pidm,robodreamer}).
In practice, all regressors are trained in a normalized action space with per-dimension training statistics $(\mu_{\mathrm{train}},\sigma_{\mathrm{train}})$ and mapped back to control units at inference.

\subsection{Overview}
\label{sec:overview}

To deal with manipulator truncation, \textit{StableIDM} employs a synergistic spatio-temporal feature refinement framework.
The framework is a three-stage system, as illustrated in Fig.~\ref{fig:architecture}. The data flow proceeds sequentially through three key stages: (i) an auxiliary robot-centric masking module, (ii) the core spatial module, Directional Feature Aggregation (DFA), and (iii) the core temporal module, Temporal Dynamics Refinement (TDR).

This interleaved architecture allows the modules to synergistically repair and refine information. Masking first cleans the input by suppressing background clutter. The temporal module then repairs the visual features \textit{before} they enter DFA, and refines the abstract features \textit{after} they exit DFA. This temporal design is deliberate: the initial temporal fusion provides a cleaner, geometrically complete spatial input for DFA, while the final TCN regressor smooths the high-level descriptor $z_t$ produced by DFA to enforce kinematic consistency.

During inference, \textit{StableIDM} runs on a fixed-length causal history window. Crucially, this window is introduced specifically to recover geometric cues lost under manipulator truncation, not to predict future frames or perform continuous  state estimation. \textit{StableIDM} functions strictly as an action decoder, outputting only the current single-step action with a fixed, estimable latency. Because it does not maintain state across windows, it avoids compounding errors over long horizons. This architecture makes the model highly suitable for real-time closed-loop control and ensures clean offline action prediction attribution. The following sections detail each component in the order of the data flow.

\subsection{Robot-centric Masking}
\label{sec:masking}

Truncation reduces the number of visible manipulator pixels and increases the proportion of background observations. Under severe truncation, high-capacity visual encoders (e.g., ViT) are prone to causal confusion, where they overfit to spurious background correlations such as textures, lighting, or stationary clutter. This causes a signal collapse problem since the model attends to irrelevant environmental distractors rather than the underlying kinematics.

To mitigate this distributional shift and preserve the action-relevant signal, we extract a robot-centric binary mask $M_t$ for each frame $I_t$ using Segment Anything~\cite{sam} as an auxiliary preprocessing step. While this is an intuitive engineering choice, it effectively filters out the background noise that dominates during severe truncation.

By applying this mask to gate the feature aggregation in the subsequent spatial and temporal modules (DFA and TDR), we provide a clean spatial input for our core architecture. This design ensures that the refinement modules allocate their representational capacity to the remaining visible geometry of the arm rather than irrelevant environmental distractors. Consequently, this masking strategy mitigates causal confusion and keeps the model focused on physical articulation, even when only a small fraction of the manipulator is visible.

\subsection{Directional Feature Aggregation}
\label{sec:DFA}

We employ a powerful pre-trained visual encoder~\cite{dinov2} as the visual backbone. We take the feature map (patch grid) $G_t \in \mathbb{R}^{C\times H\times W}$ from its final layer as the carrier of spatial information. Global self-attention in ViT tends to mix spatial information isotropically. Therefore, fine-grained directional cues are often the first to be lost during truncation or occlusion.

The purpose of DFA is to explicitly collect and amplify this direction-sensitive evidence. Because monocular RGB observations lack direct depth information, 3D structural reasoning under truncation is inherently challenging. To address this, we utilize the strong directional layout of the visible manipulator (e.g., the edges of the links) as a 2D geometric proxy. Our core insight is that the missing information from truncation is not randomly located; it lies along a geometric axis defined by the remaining visible arm segment. By extracting anisotropic features along specific angles, DFA maximizes the retention of geometric cues along the arm's orientation. This explicitly compensates for the spatial ambiguity caused by partial observability, ensuring robust geometric reasoning even under severe truncation, without relying on auxiliary depth sensors.

First, DFA forms feature components on the masked feature map using direction-aware operators at multiple analysis angles $\{\theta_k\}$. These operators are designed to be sensitive to oriented structures in the feature map, allowing them to disentangle the spatial information into its directional components. Second, it concatenates these components into an anisotropic single-frame representation. Finally, the model adaptively re-weights these angular components using global context $g_t$ extracted from the encoder's CLS and register tokens. This allows the model to dynamically emphasize directional evidence that is more reliable under the current viewpoint and truncation:

\begin{equation}
\label{eq:DFA_core}
\begin{gathered}
    z_t = \bigoplus_{k=1}^{A} \Big( w_{t,k} \big\langle \mathcal{D}_{\theta_k}(\tilde G_t)\big\rangle_{M_t} \Big), \\
    \text{where} \quad w_t = \mathrm{softmax}\big(U g_t / \tau\big).
\end{gathered}
\end{equation}

$\tilde G_t$ is the input feature map. $\mathcal{D}_{\theta_k}$ is the directional extractor, which functionally projects $G_t$ onto the $\theta_k$ angular basis. $\langle\cdot\rangle_{M_t}$ denotes masked average pooling. The global context $g_t$ is crucial: it has a global view of the scene and can infer the overall configuration and truncation state. The learned projection $U$ translates this global context into a set of directional confidences $w_t$. Intuitively, if the arm is truncated on the right, the model can learn to up-weight the $0^\circ$ extractor. 

For implementation, each masked frame is resized to $518\times 518$ and encoded into a $37\times 37$ patch grid with channel dimension $C=768$. We use $A=4$ canonical directions spanning $[0^\circ,180^\circ)$ and a lightweight projection head that maps fused features to 256 channels per direction before contextual reweighting.

Furthermore, to ensure robustness against calibration noise and minor camera displacements, we employ local view perturbations during training. The final descriptor $z_t$ is the set of these context-aware, direction-specific features. This design preserves the most action-relevant directional cues, mitigates information loss, and improves the IDM's regression stability under truncation.

\subsection{Temporal Dynamics Refinement}
\label{sec:temporal}
We recognize that single-frame spatial aggregation (DFA) alone may be insufficient during severe truncation. Therefore, we introduce a multi-stage temporal system to complete the information. This system operates synergistically at both the input visual features and output decoded features of DFA. It maintains strict causality and fixed latency.

\noindent\textbf{Temporal Fusion.}
Before DFA processes the features, a Temporal Fusion module first operates on the encoder's visual feature maps. Within our causal history window of length $K$, this fusion operates pairwise across adjacent frames. Specifically, for any given frame index $\tau \in [t-K+1, t]$ in the window, it aligns features from its preceding adjacent frame $G_{\tau-1}$ to the coordinate system of $G_{\tau}$. The features are fused via a learned visibility gate $\Pi_{\tau}$:
\begin{equation}
\label{eq:var_core}
\tilde G_{\tau} = G_{\tau} + \beta_{\mathrm{fusion}} \Pi_{\tau} \odot \big( W_{\phi}(G_{\tau-1}) - G_{\tau} \big).
\end{equation}
Here $W_{\phi}$ is a lightweight deformation and sampling field, learned by a small prediction head. $\Pi_{\tau}$ is a simultaneously predicted visibility gate, which learns to assign low confidence to regions that are truncated or have unreliable alignment. 
The fusion head is implemented as a shallow convolutional module that predicts offsets and visibility gates from adjacent-frame feature pairs.
$\beta_{\mathrm{fusion}}\!\ge\!0$ ensures this is a non-negative residual. Intuitively, temporal fusion aims to repair the input to DFA. When $G_{\tau}$ is incomplete due to severe truncation, temporal fusion borrows reliable structural information from $G_{\tau-1}$. The visibility gate $\Pi_{\tau}$ is critical, as it prevents the model from copying polluted features (e.g., background that was visible in $G_{\tau-1}$ but is now covered by the arm in $G_{\tau}$). By applying this pairwise operation sequentially, the module provides DFA with a geometrically complete spatial input sequence.

\noindent\textbf{Temporal Regressor.}
After DFA processes $\tilde G_t$ and outputs the high-level feature descriptor $z_t$, a lightweight causal temporal model operates on the feature history sequence $z_{t-K+1:t}$:
\begin{equation}
\label{eq:tcn_core}
\hat a_t = h(z_t) + \beta_{\mathrm{TCN}} f_{\mathrm{TCN}}\big(z_{t-K+1:t}\big).
\end{equation}
Here $h(\cdot)$ is a base MLP regressor from the single frame $z_t$. $f_{\mathrm{TCN}}(\cdot)$ analyzes motion trends within a fixed receptive field and predicts a residual correction. We explicitly choose a temporal model architecture that is causal and has a fixed, predictable latency. This is a critical design choice for real-time control, as it avoids the unbounded or variable latency of more complex recurrent or attention-based mechanisms.
In implementation, the temporal branch is a lightweight causal TCN with dilation factors $1,2,4,8$ over the feature history. This module's role is to refine the output of DFA by smoothing jitter and completing information that DFA could not infer alone. All components are optimized end-to-end under a unified inverse-dynamics objective.
\section{Experiments}
\label{sec:exp}

\newcommand{\best}[1]{\textbf{#1}}
\newcommand{\second}[1]{\underline{#1}}

\begin{table}[t]
    \captionsetup{justification=justified, singlelinecheck=false}
    \caption{Offline action prediction and ablation results on the AgiBot benchmark. To rigorously evaluate performance under partial observability, the dataset is divided into \textit{light} (mask ratio $>$ 15\%) and \textit{heavy} (mask ratio $<$ 15\%) truncation subsets. We report strict threshold-based accuracy (\texttt{acc}), mean per-dimension accuracy (\texttt{acc-per-dim}), and L1 prediction error (\texttt{distance}). The lower section isolates the contributions of our proposed spatio-temporal refinement modules and the robot-centric masking strategy.}
    \label{tab:light_heavy_comparison}
    \scalebox{0.85}{
    \begin{tabular}{lcccccc}
    \toprule
    \multirow{2}{*}{\textbf{Method}} &
    \multicolumn{3}{c}{\textbf{light (Truncation \textgreater 15\%)}} &
    \multicolumn{3}{c}{\textbf{heavy (Truncation \textless 15\%)}} \\
    \cmidrule(lr){2-4} \cmidrule(lr){5-7}
     & \textbf{acc$\uparrow$} & \textbf{acc-per-dim$\uparrow$} & \textbf{distance$\downarrow$}
     & \textbf{acc$\uparrow$} & \textbf{acc-per-dim$\uparrow$} & \textbf{distance$\downarrow$} \\
    \midrule
    ResNet~\cite{resnet,robodreamer} & 0.179 & 0.413 & 0.189 & 0.150 & 0.291 & 0.679 \\
    AnyPos~\cite{anypos} & 0.148 & 0.407 & 0.172 & 0.159 & 0.306 & 0.538 \\
    Vidar~\cite{vidar}  & 0.194 & 0.454 & 0.163 & 0.186 & 0.322 & 0.573 \\
    \midrule
    Ours w/o Refinement & 0.163 & 0.392 & 0.207 & 0.142 & 0.276 & 0.732 \\
    Ours w/o DFA        & 0.185 & 0.435 & 0.182 & 0.158 & 0.305 & 0.645 \\
    Ours w/o TDR        & 0.258 & 0.449 & 0.156 & 0.265 & 0.385 & 0.662 \\
    Ours w/o mask       & 0.283 & 0.461 & 0.161 & 0.288 & 0.412 & \textbf{0.475} \\
    \midrule
    \textbf{Ours (\textit{StableIDM})} & \textbf{0.286} & \textbf{0.476} & \textbf{0.142} & \textbf{0.307} & \textbf{0.447} & 0.493 \\
    \bottomrule
    \end{tabular}
    }
\end{table}

\subsection{Experimental Setup}
\label{sec:exp_setup}

We design the experiments to comprehensively validate \textit{StableIDM}. Our evaluation is structured to answer four key questions:
\begin{itemize}
    \item \textbf{Q1 (Offline Prediction Stability)}: How does \textit{StableIDM} compare to state-of-the-art baselines in offline action prediction, and does it effectively prevent prediction collapse under severe manipulator truncation?
    \item \textbf{Q2 (Real-World Replay Stability)}: Does this offline stability translate into robust and successful closed-loop control on a physical robot?
    \item \textbf{Q3 (Policy Deployment Stability)}: Can \textit{StableIDM} function as a stable executable policy for visual plans from upstream video generation models?
    \item \textbf{Q4 (Action Labeling Quality)}: Can \textit{StableIDM} function as a high-quality action annotator to improve downstream VLA model training?
\end{itemize}
Together, these questions evaluate whether improving stability under truncation is sufficient to support downstream robotic applications.

\noindent\textbf{Baselines.}
We compare \textit{StableIDM} against relevant baselines:
(1) ResNet-50, an IDM built from a standard ResNet backbone with an MLP action head following the architecture used in RoboDreamer~\cite{resnet, robodreamer}.
(2) AnyPos~\cite{anypos}, a direction-aware IDM originally proposed as a three-view model. For a fair single-view comparison, we use the official single-view variant released in the AnyPos codebase. 
(3) Vidar~\cite{vidar}, an IDM method designed for robustness to visual occlusions and clutter. It serves as a strong robustness baseline in our comparison, since it focuses on feature selection under occlusion rather than explicitly reasoning about the missing structural information of truncated manipulators.

For all experiments, the compared IDMs are trained and evaluated on the exact same data splits using identical input modalities, action parameterizations, training schedules, and loss functions whenever applicable. This rigorous evaluation protocol ensures that any observed performance differences strictly reflect the underlying model architectures rather than external training factors.

\noindent\textbf{Dataset and split details.}\label{sec:exp_dataset_index}
Our offline experiments use an AgiBot subset of 100 episodes spanning ten everyday manipulation tasks, including articulated-object interaction, tabletop pick-and-place, and cleaning-style behaviors. Each episode contains monocular RGB observations synchronized with low-level action trajectories. We use two splits, \textit{train} and \textit{eval}, shared by all methods. To improve reproducibility, we provide an episode index file (\texttt{agibot\_episodes.csv}) with \texttt{episode\_id} and \texttt{split} fields.

\subsection{Offline Action Prediction and Ablation}
\label{sec:exp_offline}

In this experiment, we evaluate \textit{StableIDM}'s action prediction ability on the AgiBot benchmark with truncation (answering Q1 from \cref{sec:exp_setup}). All methods are trained on the same AgiBot training split and evaluated on its test split using identical supervision and optimization settings.

\noindent\textbf{Experimental Setup.}
In this experiment, we use three metrics: (1) \texttt{Acc}: Overall accuracy, where a prediction is correct only if all action dimensions are within a threshold (0.1 for rotation, 0.5 for gripper). (2) \texttt{Acc-per-dim}: The average accuracy calculated per-dimension, providing a softer measure of performance. (3) \texttt{L1 Distance}: The mean L1 distance between the predicted and ground-truth action, measuring overall prediction closeness.

In order to evaluate methods' stability on truncation, we split the AgiBot benchmark into light and heavy subsets. We compute arm pixel occupancy using the arm mask and define it as the ratio of masked arm pixels to all image pixels. On AgiBot, this occupancy typically falls in the 10\% to 20\% range. Therefore, we set 15\% as a data-centered threshold to split light and heavy truncation, ensuring a reproducible and non-arbitrary evaluation. Data is classified as less truncation (light) if its manipulator ratio is higher than 15\%, and as severe truncation (heavy) otherwise. Details on mask construction and quality analysis are provided in \cref{sec:exp_mask_quality}.

We provide data sources and a list of used samples to facilitate reproduction (\cref{sec:exp_dataset_index}). We also analyze segmentation quality in \cref{sec:exp_mask_quality} and observe that \textit{StableIDM} is robust to moderate segmentation noise during real-world deployments.

\noindent\textbf{Results Analysis.}
\cref{tab:light_heavy_comparison} shows the offline prediction results under different levels of manipulator truncation. Compared with the light subset, the heavy subset leads to a notable degradation for existing IDMs, with accuracy rates dropping and prediction errors increasing substantially when the manipulator is heavily truncated. \textit{StableIDM} is markedly more robust in this regime, retaining high per-dimension accuracy and achieving the lowest prediction error, indicating stronger tolerance to missing or truncated manipulator observations.

The critical regime is the heavy (severe truncation) subset. Here, the performance of all baselines deteriorates sharply, whereas full \textit{StableIDM} maintains high performance and achieves the best overall results. On the heavy subset, its \texttt{acc} improves from 18.6\% to 30.7\%, \texttt{acc-per-dim} improves from 32.2\% to 44.7\%, and the \texttt{distance} is reduced from 0.573 to 0.493. This demonstrates that our spatio-temporal feature refinement framework effectively compensates for lost information and maintains stable, accurate predictions where other IDMs fail.

\noindent\textbf{Ablation Study.}
\cref{tab:light_heavy_comparison} also ablates our model's components to validate their complementary utility.
(1) \textbf{Ours w/o Refinement}: Removing the entire spatio-temporal refinement stack causes performance to drop notably across all metrics, supporting its foundational importance for stability under truncation.
(2) \textbf{Ours w/o DFA}: Removing the Directional Feature Aggregation module leads to a sharp accuracy drop under heavy truncation. This confirms the importance of DFA for spatial localization and structural reasoning when visual evidence is scarce.
(3) \textbf{Ours w/o TDR}: Removing the Temporal Dynamics Refinement module causes a severe spike in L1 distance despite relatively preserved accuracy. This discrepancy indicates temporal jitter and confirms the necessity of TDR for kinematic smoothing.
(4) \textbf{Ours w/o mask}: Removing the robot-centric masking causes the model to overfit to static background cues. While exploiting this spurious shortcut can superficially reduce the mean L1 distance under heavy truncation (0.475 vs. 0.493) by anchoring predictions to rigid scene elements, it causes a severe drop in the strict success metric (\texttt{acc} drops to 28.8\%). This discrepancy highlights that relying on background textures fails to capture the true kinematic articulation required for successful execution.

\noindent\textbf{Mask-quality robustness.}\label{sec:exp_mask_quality}
To directly assess sensitivity to segmentation quality, we evaluate a dedicated subset with severe mask failures (e.g., large missing manipulator regions or strong background leakage), and compare it to the full evaluation set.

\begin{table}[t]
    \captionsetup{justification=justified, singlelinecheck=false}
    \centering
    \caption{Effect of mask quality on mean L1 error. Lower is better.}
    \label{tab:mask-quality}
    \begin{tabular}{lcc}
    \toprule
    Split & Severe Mask Failures & All Episodes \\
    \midrule
    Light truncation  & 0.149 & 0.142 \\
    Heavy truncation  & 0.517 & 0.493 \\
    \bottomrule
    \end{tabular}
\end{table}

As expected, the severe-failure subset is harder in both truncation regimes, but the increase remains moderate (0.149 vs 0.142 under light truncation; 0.517 vs 0.493 under heavy truncation), indicating that \textit{StableIDM} remains stable under mask degradation.

\begin{figure*}[h]
  \centering
  \captionsetup{justification=justified, singlelinecheck=false}
  \includegraphics[width=\linewidth]{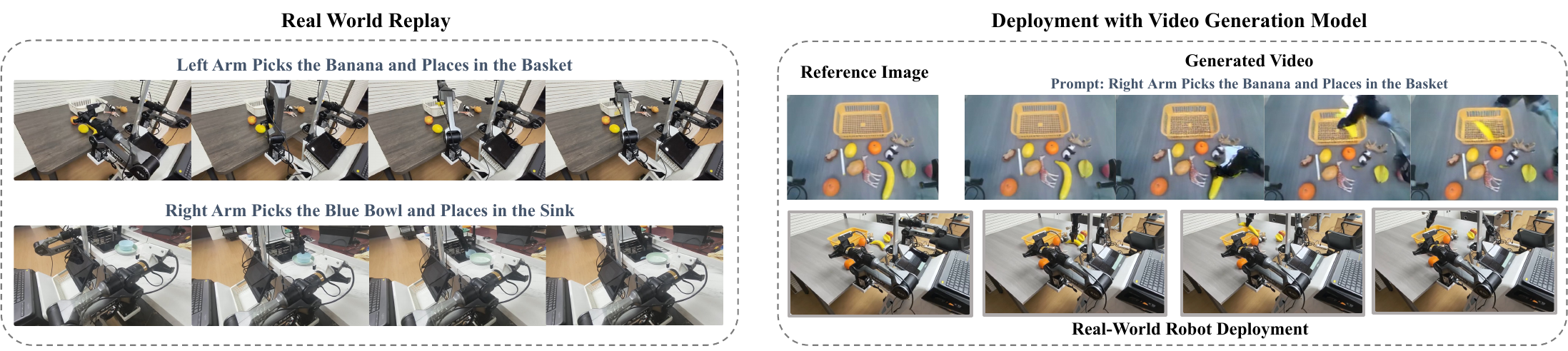}
  \caption{Experimental visualizations. (a) \textbf{Real-World Replay}: This part includes representative pick-and-place tasks: Pick up a banana and a blue bowl and place them in designated locations. Actions decoded by \textit{StableIDM} execute smoothly and stably in real world. (b) \textbf{Video-Plan Deployment}: Given an initial frame and a text prompt, a video plan is generated by a video generation model. Then \textit{StableIDM} decodes actions from the generated video and replays them in the real world, yielding successful executions and indicating a stable policy when paired with video generation models.}
  \label{fig:exp_figure}
\end{figure*}

\subsection{Real-World Replay on the Physical Platform} 
\label{sec:exp_replay}

Offline metrics do not fully capture the challenges of real-world execution, where physical dynamics and sensor noise can introduce additional instability. Therefore, we investigate if this superior offline performance translates to real-world stability (answering Q2 from \cref{sec:exp_setup}). 

\noindent\textbf{Experimental Setup.}
To validate the accuracy of the predicted actions, we generate action sequences from our real-world test set videos and execute them in open-loop replay on our physical manipulation platform.
We deliberately adopt open-loop rather than closed-loop execution to strictly isolate the model's inherent action decoding fidelity, preventing continuous visual feedback from masking instantaneous prediction errors.
Crucially, to thoroughly evaluate the model's generalization, our evaluation goes beyond standard pick-and-place tasks. It encompasses significantly more challenging scenarios, including Microwave Operation (which involves manipulating articulated doors) and Sink Cleaning (which tackles severe occlusion within recessed areas). 

Each sequence is replayed multiple times to account for stochasticity. To rigorously evaluate performance across these diverse scenarios, we define success as the strict completion of the entire task sequence in a single trial. For Pick \& Place, the robot must successfully grasp and transport the object to the target. For Microwave Operation, it requires properly interacting with the appliance (e.g., articulating the door or placing an object inside) without severe collisions. For Sink Cleaning, it involves accurate grasping and placement despite severe partial occlusions in the recessed area. All IDMs use the same training settings and are executed with the same control frequency on the physical platform.

\noindent\textbf{Results Analysis.}
\cref{tab:replay} demonstrates a clear advantage for \textit{StableIDM}. Baselines struggle to achieve reliable success, with the strongest, Vidar, averaging only 37.7\%. They also lack consistency across complex tasks (e.g., Vidar drops to 29.1\% on Sink Cleaning). Conversely, \textit{StableIDM} consistently outperforms all baselines, attaining 53.8\% on Pick \& Place, and remarkably maintaining 42.3\% and 46.2\% on the constrained Microwave and Sink Cleaning tasks, averaging 47.4\%. This confirms that our offline stability (\cref{sec:exp_offline}) successfully translates into robust real-world physical execution.

\begin{table}[t]
    \centering
    \captionsetup{justification=justified, singlelinecheck=false}
    \caption{Real-world open-loop replay success rates (\%) across diverse physical manipulation tasks. To strictly isolate the inherent action decoding fidelity of each model and prevent visual feedback from masking prediction errors, all synthetic sequences are executed entirely open-loop at a fixed control frequency.}
    \label{tab:replay}
    \setlength{\tabcolsep}{4pt} 
    \begin{tabular}{lcccc}
    \toprule
    \textbf{Method} & \textbf{Pick \& Place} & \textbf{Microwave} & \textbf{Sink Cleaning} & \textbf{Average} \\
    \midrule
    ResNet & 23.1 & 27.4 & 19.7 & 23.4 \\
    AnyPos & 34.6 & 39.2 & 34.5 & 36.1 \\
    Vidar  & 46.2 & 37.9 & 29.1 & 37.7 \\
    \midrule
    \textbf{Ours (\textit{StableIDM})} & \textbf{53.8} & \textbf{42.3} & \textbf{46.2} & \textbf{47.4} \\
    \bottomrule
    \end{tabular}
\end{table}

\subsection{Video-Plan Deployment with Video Generation}
\label{sec:exp_policy}

We evaluate \textit{StableIDM} as an executable policy when combined with a video generation model (answering Q3 from \cref{sec:exp_setup}). In this setting, the IDM must decode actions from synthetic visual plans produced by a monocular generator, which differ from real camera observations and provide limited depth cues.

\noindent\textbf{Experimental Setup.}
We use a single, fixed video generation model to create visual plans for various tasks. All methods decode plans produced by the same pretrained monocular video generator given identical initial frames and prompts. We then employ each baseline and \textit{StableIDM} to convert these visual plans into executable actions under identical open-loop test conditions. 

In monocular video generation settings, visual artifacts such as spatial interpenetration are common, which amplifies contact and closure ambiguity. Despite these challenges, we rigorously evaluate end-to-end task performance by reporting \textit{Grasp Success}. While recent methods like SuSIE and CLOVER improve grasping via closed-loop correction, we strictly maintain an open-loop evaluation to isolate the inherent robustness of the action decoders.

\begin{table}[t]
    \centering
    \captionsetup{justification=justified, singlelinecheck=false}
    \caption{Video-plan deployment evaluation. We report the end-to-end grasp success rates (\%) when utilizing different IDMs to decode synthetic visual plans. All baseline methods and \textit{StableIDM} process identical outputs from a shared monocular video generator and apply the exact same action normalization procedures.}
    \label{tab:policy}
    \begin{tabular}{lc}
    \toprule
    \textbf{Method}  & \textbf{Grasp Success (\%)} \\
    \midrule
    ResNet   & 11.5 \\
    Vidar    & 26.9 \\
    AnyPos   & 42.3 \\
    \midrule
    \textbf{Ours (\textit{StableIDM})} & \textbf{53.8} \\
    \bottomrule
    \end{tabular}
\end{table}

\noindent\textbf{Results Analysis.}
As reported in \cref{tab:policy}, \textit{StableIDM} demonstrates a clear advantage in this generative pipeline. While inherent generation artifacts cap the absolute performance across all methods, our model achieves a 53.8\% grasp success rate, substantially outperforming the strongest baseline, AnyPos (42.3\%).

Although visually plausible, synthetic video plans introduce severe domain shifts beyond standard truncation. Baselines lacking explicit spatio-temporal refinement are highly susceptible to these shifts, often overfitting to generative artifacts and yielding erratic actions. In contrast, the spatio-temporal completion framework of \textit{StableIDM} provides a strong prior over coherent dynamics. This enables the model to effectively smooth over minor artifacts and temporal inconsistencies, maintaining stable predictions from imperfect generative inputs. These findings confirm \textit{StableIDM} as a highly reliable policy decoder for generate-decode-execute robotic pipelines.

\begin{figure*}[htbp]
    \centering
    \captionsetup{justification=justified, singlelinecheck=false}
    \includegraphics[width=\linewidth]{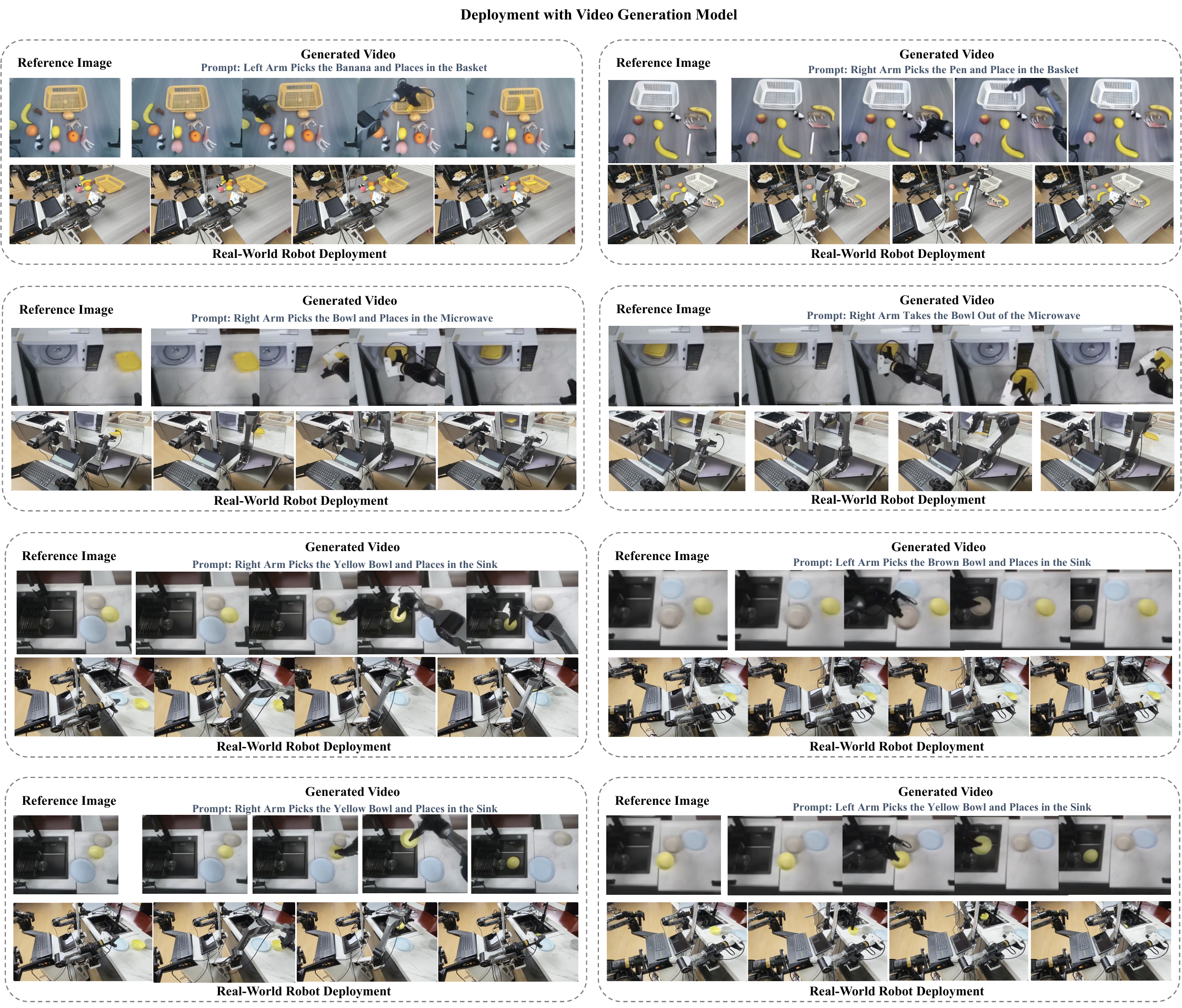}
    \caption{Deployment of \textit{StableIDM} with a video generation model. Each panel shows frames from a video generated from a language prompt together with frames from the corresponding real robot execution under the same prompt. The first row corresponds to the pick and place task and contains two prompts in which the robot moves tabletop objects into a basket. The second row corresponds to the microwave operation task and contains two prompts in which the robot moves a bowl between the workspace and the interior of the microwave. The third row corresponds to the sink cleaning task and contains four prompts in which the robot moves bowls from the table into the sink region using either the left or the right arm. Across these eight prompts the figure illustrates how a single inverse dynamics model trained on real videos can decode synthetic videos into real robot actions across different tasks and truncation patterns.}
    \label{fig:deployment}
\end{figure*}

The qualitative deployment cases cover pick and place, microwave operation, and sink cleaning. Across these scenarios, generated videos often contain truncation and minor visual artifacts, while decoded executions remain coherent and task-consistent, matching the quantitative trend in \cref{tab:policy}.

\subsection{IDM-Labeled Videos for VLA Training}
\label{sec:exp_vla}

In this experiment, we validate \textit{StableIDM}'s second key role (answering Q4 from \cref{sec:exp_setup}): as a high-fidelity action annotator for downstream VLA models.

The preceding experiments have confirmed that \textit{StableIDM} is a robust action decoder, capable of handling both real-world truncation (\cref{sec:exp_replay}) and noisy generated plans (\cref{sec:exp_policy}). We now investigate whether this stability of \textit{StableIDM} is sufficient to generate useful action labels for training VLA models. This experiment targets a practical challenge: VLA training is often bottlenecked by the reliance on expensive and scarce human-annotated data.
We hypothesize that \textit{StableIDM} can function as a data engine, leveraging its stable predictions to automatically label generated videos and thus effectively augment real datasets.

\noindent\textbf{Experimental Setup.}
We adopt a standard VLA model Pi-0.5 and train it under two key data regimes in \cref{tab:vla}:
Real Only (Baseline): Trained using only the real human-annotated data.
Real + Generated (\textit{StableIDM}): Trained using the complete real dataset, augmented with generated videos automatically labeled by our \textit{StableIDM}.
We keep the model architecture and training hyperparameters identical across all data regimes.

\begin{table}[h]
    \centering
    \captionsetup{justification=justified, singlelinecheck=false}
    \caption{Downstream task success rates (\%) for VLA training. We evaluate a standard Pi-0.5 VLA model trained under two distinct data regimes: a baseline using strictly human-annotated real data, and an augmented setting that incorporates synthetic videos automatically labeled by our \textit{StableIDM}.}
    \label{tab:vla}
    \begin{tabular}{lc}
    \toprule
    Training Data Regime & Success Rate (\%) \\
    \midrule
    Real Only (Baseline) & 35.3 \\
    Real + Generated (\textit{StableIDM}) & \textbf{52.9} \\
    \bottomrule
    \end{tabular}
\end{table}

\noindent\textbf{Results Analysis.}
As shown in \cref{tab:vla}, the baseline model trained only on real data achieves a 35.3\% success rate, while augmenting with \textit{StableIDM} labeled generated videos increases performance to 52.9\%, suggesting the generated labels provide a useful training signal for VLAs.

These results complement our previous experiments: in addition to improving offline metrics in \cref{sec:exp_offline}, real-world replay in \cref{sec:exp_replay}, and policy execution in \cref{sec:exp_policy}, \textit{StableIDM} can also be used to generate labels that improve downstream VLA training. This indicates that the stability and accuracy of \textit{StableIDM} are sufficient for it to serve as a practical data annotation engine, helping VLA training scale beyond purely human collected datasets.
\section{Conclusion}
\label{sec:conclusion}

We presented \textit{StableIDM}, a robust inverse dynamics model that decodes stable actions under severe manipulator truncation. By integrating robot-centric masking, Directional Feature Aggregation (DFA), and Temporal Dynamics Refinement (TDR), the framework effectively recovers missing structural information from partial observations and prevents error accumulation.

Extensive evaluations validate its versatility across four key settings: \textit{StableIDM} significantly improves offline prediction accuracy under severe truncation by 12.1\% and translates this stability into robust physical execution, increasing average task success by 9.7\% across geometrically complex real-world tasks. Furthermore, it excels as both a reliable deployable policy, boosting grasp success rates by 11.5\% on video-generated plans, and a scalable offline annotator that improves downstream VLA training by 17.6\%. By bridging the critical gap between high-level visual generation and low-level physical control, our approach unlocks the potential of generative foundation models in robotics.
At the same time, \textit{StableIDM} still has two limitations. First, the current framework mainly targets single-task execution, and extending it to long-horizon multi-step manipulation with stronger sequential reasoning remains open. Second, the video generation model and IDM are trained in a decoupled manner, which can introduce representation mismatch between synthetic observations and action decoding dynamics. Future work will therefore focus on multi-view extensions for stronger geometric observability and joint optimization of generation and inverse-dynamics modules to improve alignment and reduce execution drift in complex scenarios.

\clearpage
\setcitestyle{numbers}
\bibliographystyle{plainnat}
\bibliography{main}

@article{anypos,
  title={AnyPos: Automated Task-Agnostic Actions for Bimanual Manipulation},
  author={Tan, Hengkai and Feng, Yao and Mao, Xinyi and Huang, Shuhe and Liu, Guodong and Hao, Zhongkai and Su, Hang and Zhu, Jun},
  journal={arXiv preprint arXiv:2507.12768},
  year={2025}
}

@inproceeding{octo,
  title={Octo: An open-source generalist robot policy},
  author={Mees, Oier and Ghosh, Dibya and Pertsch, Karl and Black, Kevin and Walke, Homer Rich and Dasari, Sudeep and Hejna, Joey and Kreiman, Tobias and Xu, Charles and Luo, Jianlan and others},
  booktitle={First Workshop on Vision-Language Models for Navigation and Manipulation at ICRA 2024},
  year={2024}
}

@article{rdt,
  title={Rdt-1b: a diffusion foundation model for bimanual manipulation},
  author={Liu, Songming and Wu, Lingxuan and Li, Bangguo and Tan, Hengkai and Chen, Huayu and Wang, Zhengyi and Xu, Ke and Su, Hang and Zhu, Jun},
  journal={arXiv preprint arXiv:2410.07864},
  year={2024}
}

@article{openvla,
  title={Openvla: An open-source vision-language-action model},
  author={Kim, Moo Jin and Pertsch, Karl and Karamcheti, Siddharth and Xiao, Ted and Balakrishna, Ashwin and Nair, Suraj and Rafailov, Rafael and Foster, Ethan and Lam, Grace and Sanketi, Pannag and others},
  journal={arXiv preprint arXiv:2406.09246},
  year={2024}
}

@article{bridge,
  title={Bridge data: Boosting generalization of robotic skills with cross-domain datasets},
  author={Ebert, Frederik and Yang, Yanlai and Schmeckpeper, Karl and Bucher, Bernadette and Georgakis, Georgios and Daniilidis, Kostas and Finn, Chelsea and Levine, Sergey},
  journal={arXiv preprint arXiv:2109.13396},
  year={2021}
}

@article{diffusionpolicy,
  title={Diffusion policy: Visuomotor policy learning via action diffusion},
  author={Chi, Cheng and Xu, Zhenjia and Feng, Siyuan and Cousineau, Eric and Du, Yilun and Burchfiel, Benjamin and Tedrake, Russ and Song, Shuran},
  journal={The International Journal of Robotics Research},
  year={2023},
}

@article{susie,
  title={Zero-shot robotic manipulation with pretrained image-editing diffusion models},
  author={Black, Kevin and Nakamoto, Mitsuhiko and Atreya, Pranav and Walke, Homer and Finn, Chelsea and Kumar, Aviral and Levine, Sergey},
  journal={arXiv preprint arXiv:2310.10639},
  year={2023}
}

@article{unipi,
  title={Learning universal policies via text-guided video generation},
  author={Du, Yilun and Yang, Sherry and Dai, Bo and Dai, Hanjun and Nachum, Ofir and Tenenbaum, Josh and Schuurmans, Dale and Abbeel, Pieter},
  journal={Advances in neural information processing systems},
  year={2023}
}

@article{unisim,
  title={3d diffusion policy: Generalizable visuomotor policy learning via simple 3d representations},
  author={Ze, Yanjie and Zhang, Gu and Zhang, Kangning and Hu, Chenyuan and Wang, Muhan and Xu, Huazhe},
  journal={arXiv preprint arXiv:2403.03954},
  year={2024}
}

@article{robodreamer,
  title={Robodreamer: Learning compositional world models for robot imagination},
  author={Zhou, Siyuan and Du, Yilun and Chen, Jiaben and Li, Yandong and Yeung, Dit-Yan and Gan, Chuang},
  journal={arXiv preprint arXiv:2404.12377},
  year={2024}
}

@article{dinov2,
  title={Dinov2: Learning robust visual features without supervision},
  author={Oquab, Maxime and Darcet, Timoth{\'e}e and Moutakanni, Th{\'e}o and Vo, Huy and Szafraniec, Marc and Khalidov, Vasil and Fernandez, Pierre and Haziza, Daniel and Massa, Francisco and El-Nouby, Alaaeldin and others},
  journal={arXiv preprint arXiv:2304.07193},
  year={2023}
}

@article{vit,
  title={An image is worth 16x16 words: Transformers for image recognition at scale},
  author={Dosovitskiy, Alexey and Beyer, Lucas and Kolesnikov, Alexander and Weissenborn, Dirk and Zhai, Xiaohua and Unterthiner, Thomas and Dehghani, Mostafa and Minderer, Matthias and Heigold, Georg and Gelly, Sylvain and others},
  journal={arXiv preprint arXiv:2010.11929},
  year={2020}
}

@article{agibot,
  title={Agibot world colosseo: A large-scale manipulation platform for scalable and intelligent embodied systems},
  author={Bu, Qingwen and Cai, Jisong and Chen, Li and Cui, Xiuqi and Ding, Yan and Feng, Siyuan and Gao, Shenyuan and He, Xindong and Hu, Xuan and Huang, Xu and others},
  journal={arXiv preprint arXiv:2503.06669},
  year={2025}
}

@inproceedings{resnet,
  title={Deep residual learning for image recognition},
  author={He, Kaiming and Zhang, Xiangyu and Ren, Shaoqing and Sun, Jian},
  booktitle={Proceedings of the IEEE conference on computer vision and pattern recognition},
  year={2016}
}

@article{pidm,
  title={Predictive inverse dynamics models are scalable learners for robotic manipulation},
  author={Tian, Yang and Yang, Sizhe and Zeng, Jia and Wang, Ping and Lin, Dahua and Dong, Hao and Pang, Jiangmiao},
  journal={arXiv preprint arXiv:2412.15109},
  year={2024}
}

@article{vidman,
  title={Vidman: Exploiting implicit dynamics from video diffusion model for effective robot manipulation},
  author={Wen, Youpeng and Lin, Junfan and Zhu, Yi and Han, Jianhua and Xu, Hang and Zhao, Shen and Liang, Xiaodan},
  journal={Advances in Neural Information Processing Systems},
  year={2024}
}

@article{dreamgen,
  title={DreamGen: Unlocking Generalization in Robot Learning through Neural Trajectories},
  author={Jang, Joel and Ye, Seonghyeon and Lin, Zongyu and Xiang, Jiannan and Bjorck, Johan and Fang, Yu and Hu, Fengyuan and Huang, Spencer and Kundalia, Kaushil and Lin, Yen-Chen and others},
  journal={arXiv e-prints},
  year={2025}
}

@article{vidar,
  title={Vidar: Embodied Video Diffusion Model for Generalist Bimanual Manipulation},
  author={Feng, Yao and Tan, Hengkai and Mao, Xinyi and Liu, Guodong and Huang, Shuhe and Xiang, Chendong and Su, Hang and Zhu, Jun},
  journal={arXiv preprint arXiv:2507.12898},
  year={2025}
}

@article{vpp,
  title={Video prediction policy: A generalist robot policy with predictive visual representations},
  author={Hu, Yucheng and Guo, Yanjiang and Wang, Pengchao and Chen, Xiaoyu and Wang, Yen-Jen and Zhang, Jianke and Sreenath, Koushil and Lu, Chaochao and Chen, Jianyu},
  journal={arXiv preprint arXiv:2412.14803},
  year={2024}
}

@article{aloha1,
  title={Learning fine-grained bimanual manipulation with low-cost hardware},
  author={Zhao, Tony Z and Kumar, Vikash and Levine, Sergey and Finn, Chelsea},
  journal={arXiv preprint arXiv:2304.13705},
  year={2023}
}

@article{aloha2,
  title={Aloha unleashed: A simple recipe for robot dexterity},
  author={Zhao, Tony Z and Tompson, Jonathan and Driess, Danny and Florence, Pete and Ghasemipour, Kamyar and Finn, Chelsea and Wahid, Ayzaan},
  journal={arXiv preprint arXiv:2410.13126},
  year={2024}
}

@inproceedings{robotwin1,
  title={Robotwin: Dual-arm robot benchmark with generative digital twins (early version)},
  author={Mu, Yao and Chen, Tianxing and Peng, Shijia and Chen, Zanxin and Gao, Zeyu and Zou, Yude and Lin, Lunkai and Xie, Zhiqiang and Luo, Ping},
  booktitle={European Conference on Computer Vision},
  year={2024},
}

@article{robotwin2,
  title={Robotwin 2.0: A scalable data generator and benchmark with strong domain randomization for robust bimanual robotic manipulation},
  author={Chen, Tianxing and Chen, Zanxin and Chen, Baijun and Cai, Zijian and Liu, Yibin and Li, Zixuan and Liang, Qiwei and Lin, Xianliang and Ge, Yiheng and Gu, Zhenyu and others},
  journal={arXiv preprint arXiv:2506.18088},
  year={2025}
}

@article{manibox,
  title={Manibox: Enhancing spatial grasping generalization via scalable simulation data generation},
  author={Tan, Hengkai and Xu, Xuezhou and Ying, Chengyang and Mao, Xinyi and Liu, Songming and Zhang, Xingxing and Su, Hang and Zhu, Jun},
  journal={arXiv preprint arXiv:2411.01850},
  year={2024}
}

@article{sora,
  title={Sora: A review on background, technology, limitations, and opportunities of large vision models},
  author={Liu, Yixin and Zhang, Kai and Li, Yuan and Yan, Zhiling and Gao, Chujie and Chen, Ruoxi and Yuan, Zhengqing and Huang, Yue and Sun, Hanchi and Gao, Jianfeng and others},
  journal={arXiv preprint arXiv:2402.17177},
  year={2024}
}

@article{svd,
  title={Stable video diffusion: Scaling latent video diffusion models to large datasets},
  author={Blattmann, Andreas and Dockhorn, Tim and Kulal, Sumith and Mendelevitch, Daniel and Kilian, Maciej and Lorenz, Dominik and Levi, Yam and English, Zion and Voleti, Vikram and Letts, Adam and others},
  journal={arXiv preprint arXiv:2311.15127},
  year={2023}
}

@inproceedings{latentVideoDiffusion,
  title={Align your latents: High-resolution video synthesis with latent diffusion models},
  author={Blattmann, Andreas and Rombach, Robin and Ling, Huan and Dockhorn, Tim and Kim, Seung Wook and Fidler, Sanja and Kreis, Karsten},
  booktitle={Proceedings of the IEEE/CVF conference on computer vision and pattern recognition},
  pages={22563--22575},
  year={2023}
}

@article{wan,
  title={Wan: Open and advanced large-scale video generative models},
  author={Wan, Team and Wang, Ang and Ai, Baole and Wen, Bin and Mao, Chaojie and Xie, Chen-Wei and Chen, Di and Yu, Feiwu and Zhao, Haiming and Yang, Jianxiao and others},
  journal={arXiv preprint arXiv:2503.20314},
  year={2025}
}

@article{cogvideo,
  title={Cogvideo: Large-scale pretraining for text-to-video generation via transformers},
  author={Hong, Wenyi and Ding, Ming and Zheng, Wendi and Liu, Xinghan and Tang, Jie},
  journal={arXiv preprint arXiv:2205.15868},
  year={2022}
}

@article{cogvideox,
  title={Cogvideox: Text-to-video diffusion models with an expert transformer},
  author={Yang, Zhuoyi and Teng, Jiayan and Zheng, Wendi and Ding, Ming and Huang, Shiyu and Xu, Jiazheng and Yang, Yuanming and Hong, Wenyi and Zhang, Xiaohan and Feng, Guanyu and others},
  journal={arXiv preprint arXiv:2408.06072},
  year={2024}
}

@article{cosmos1,
  title={Cosmos world foundation model platform for physical ai},
  author={Agarwal, Niket and Ali, Arslan and Bala, Maciej and Balaji, Yogesh and Barker, Erik and Cai, Tiffany and Chattopadhyay, Prithvijit and Chen, Yongxin and Cui, Yin and Ding, Yifan and others},
  journal={arXiv preprint arXiv:2501.03575},
  year={2025}
}

@article{cosmos2,
  title={Cosmos-reason1: From physical common sense to embodied reasoning},
  author={Azzolini, Alisson and Bai, Junjie and Brandon, Hannah and Cao, Jiaxin and Chattopadhyay, Prithvijit and Chen, Huayu and Chu, Jinju and Cui, Yin and Diamond, Jenna and Ding, Yifan and others},
  journal={arXiv preprint arXiv:2503.15558},
  year={2025}
}

@article{robomind,
  title={Robomind: Benchmark on multi-embodiment intelligence normative data for robot manipulation},
  author={Wu, Kun and Hou, Chengkai and Liu, Jiaming and Che, Zhengping and Ju, Xiaozhu and Yang, Zhuqin and Li, Meng and Zhao, Yinuo and Xu, Zhiyuan and Yang, Guang and others},
  journal={arXiv preprint arXiv:2412.13877},
  year={2024}
}

@inproceedings{sam,
  title={Segment anything},
  author={Kirillov, Alexander and Mintun, Eric and Ravi, Nikhila and Mao, Hanzi and Rolland, Chloe and Gustafson, Laura and Xiao, Tete and Whitehead, Spencer and Berg, Alexander C and Lo, Wan-Yen and others},
  booktitle={Proceedings of the IEEE/CVF international conference on computer vision},
  year={2023}
}

@article{uva,
  title={Unified video action model},
  author={Li, Shuang and Gao, Yihuai and Sadigh, Dorsa and Song, Shuran},
  journal={arXiv preprint arXiv:2503.00200},
  year={2025}
}

@inproceedings{openxembodiment,
  title={Open x-embodiment: Robotic learning datasets and rt-x models: Open x-embodiment collaboration 0},
  author={O’Neill, Abby and Rehman, Abdul and Maddukuri, Abhiram and Gupta, Abhishek and Padalkar, Abhishek and Lee, Abraham and Pooley, Acorn and Gupta, Agrim and Mandlekar, Ajay and Jain, Ajinkya and others},
  booktitle={2024 IEEE International Conference on Robotics and Automation (ICRA)},
  organization={IEEE}
}

@article{gr2,
  title={Gr-2: A generative video-language-action model with web-scale knowledge for robot manipulation},
  author={Cheang, Chi-Lam and Chen, Guangzeng and Jing, Ya and Kong, Tao and Li, Hang and Li, Yifeng and Liu, Yuxiao and Wu, Hongtao and Xu, Jiafeng and Yang, Yichu and others},
  journal={arXiv preprint arXiv:2410.06158},
  year={2024}
}

@article{gen2act,
  title={Gen2act: Human video generation in novel scenarios enables generalizable robot manipulation},
  author={Bharadhwaj, Homanga and Dwibedi, Debidatta and Gupta, Abhinav and Tulsiani, Shubham and Doersch, Carl and Xiao, Ted and Shah, Dhruv and Xia, Fei and Sadigh, Dorsa and Kirmani, Sean},
  journal={arXiv preprint arXiv:2409.16283},
  year={2024}
}

@article{wow,
  title={Wow: Towards a world omniscient world model through embodied interaction},
  author={Chi, Xiaowei and Jia, Peidong and Fan, Chun-Kai and Ju, Xiaozhu and Mi, Weishi and Zhang, Kevin and Qin, Zhiyuan and Tian, Wanxin and Ge, Kuangzhi and Li, Hao and others},
  journal={arXiv preprint arXiv:2509.22642},
  year={2025}
}

@article{transformer,
  title={Attention is all you need},
  author={Vaswani, Ashish and Shazeer, Noam and Parmar, Niki and Uszkoreit, Jakob and Jones, Llion and Gomez, Aidan N and Kaiser, {\L}ukasz and Polosukhin, Illia},
  journal={Advances in neural information processing systems},
  year={2017}
}

@article{unleashing,
  title={Unleashing large-scale video generative pre-training for visual robot manipulation},
  author={Wu, Hongtao and Jing, Ya and Cheang, Chilam and Chen, Guangzeng and Xu, Jiafeng and Li, Xinghang and Liu, Minghuan and Li, Hang and Kong, Tao},
  journal={arXiv preprint arXiv:2312.13139},
  year={2023}
}

@article{hunyuanvideo,
  title={Hunyuanvideo: A systematic framework for large video generative models},
  author={Kong, Weijie and Tian, Qi and Zhang, Zijian and Min, Rox and Dai, Zuozhuo and Zhou, Jin and Xiong, Jiangfeng and Li, Xin and Wu, Bo and Zhang, Jianwei and others},
  journal={arXiv preprint arXiv:2412.03603},
  year={2024}
}

@article{pi0,
  title={Pi0: A Vision-Language-Action Flow Model for General Robot Control},
  author={Black, Kevin and Brown, Noah and Driess, Danny and Esmail, Adnan and Equi, Michael and Finn, Chelsea and Fusai, Niccolo and Groom, Lachy and Hausman, Karol and Ichter, Brian and others},
  journal={arXiv preprint arXiv:2410.24164},
  year={2024}
}

@article{pi0.5,
  title={Pi0.5: a Vision-Language-Action Model with Open-World Generalization},
  author={Intelligence, Physical and Black, Kevin and Brown, Noah and Darpinian, James and Dhabalia, Karan and Driess, Danny and Esmail, Adnan and Equi, Michael and Finn, Chelsea and Fusai, Niccolo and others},
  journal={arXiv preprint arXiv:2504.16054},
  year={2025}
}

@article{Vidu,
  title={Vidu: a highly consistent, dynamic and skilled text-to-video generator with diffusion models},
  author={Bao, Fan and Xiang, Chendong and Yue, Gang and He, Guande and Zhu, Hongzhou and Zheng, Kaiwen and Zhao, Min and Liu, Shilong and Wang, Yaole and Zhu, Jun},
  journal={arXiv preprint arXiv:2405.04233},
  year={2024}
}

@inproceedings{genie,
  title={Genie: Generative interactive environments},
  author={Bruce, Jake and Dennis, Michael D and Edwards, Ashley and Parker-Holder, Jack and Shi, Yuge and Hughes, Edward and Lai, Matthew and Mavalankar, Aditi and Steigerwald, Richie and Apps, Chris and others},
  booktitle={Forty-first International Conference on Machine Learning},
  year={2024}
}

@article{bu2024closed,
  title={Closed-loop visuomotor control with generative expectation for robotic manipulation},
  author={Bu, Qingwen and Zeng, Jia and Chen, Li and Yang, Yanchao and Zhou, Guyue and Yan, Junchi and Luo, Ping and Cui, Heming and Ma, Yi and Li, Hongyang},
  journal={Advances in Neural Information Processing Systems},
  year={2024}
}

@article{hoque2025egodex,
  title={EgoDex: Learning Dexterous Manipulation from Large-Scale Egocentric Video},
  author={Hoque, Ryan and Huang, Peide and Yoon, David J and Sivapurapu, Mouli and Zhang, Jian},
  journal={arXiv preprint arXiv:2505.11709},
  year={2025}
}

@article{yuan2025roboengine,
  title={RoboEngine: Plug-and-Play Robot Data Augmentation with Semantic Robot Segmentation and Background Generation},
  author={Yuan, Chengbo and Joshi, Suraj and Zhu, Shaoting and Su, Hang and Zhao, Hang and Gao, Yang},
  journal={arXiv preprint arXiv:2503.18738},
  year={2025}
}

@article{ACT,
  title={Mobile aloha: Learning bimanual mobile manipulation with low-cost whole-body teleoperation},
  author={Fu, Zipeng and Zhao, Tony Z and Finn, Chelsea},
  journal={arXiv preprint arXiv:2401.02117},
  year={2024}
}

@article{diffusionpolicy2,
  title={Diffusion policy policy optimization},
  author={Ren, Allen Z and Lidard, Justin and Ankile, Lars L and Simeonov, Anthony and Agrawal, Pulkit and Majumdar, Anirudha and Burchfiel, Benjamin and Dai, Hongkai and Simchowitz, Max},
  journal={arXiv preprint arXiv:2409.00588},
  year={2024}
}

@article{gigabrain0,
  title={GigaBrain-0: A World Model-Powered Vision-Language-Action Model},
  author={Team, GigaBrain and Ye, Angen and Wang, Boyuan and Ni, Chaojun and Huang, Guan and Zhao, Guosheng and Li, Haoyun and Li, Jie and Zhu, Jiagang and Feng, Lv and others},
  journal={arXiv preprint arXiv:2510.19430},
  year={2025}
}

@article{walloss,
  title={Igniting vlms toward the embodied space},
  author={Zhai, Andy and Liu, Brae and Fang, Bruno and Cai, Chalse and Ma, Ellie and Yin, Ethan and Wang, Hao and Zhou, Hugo and Wang, James and Shi, Lights and others},
  journal={arXiv preprint arXiv:2509.11766},
  year={2025}
}

@article{g0,
  title={Galaxea open-world dataset and g0 dual-system vla model},
  author={Jiang, Tao and Yuan, Tianyuan and Liu, Yicheng and Lu, Chenhao and Cui, Jianning and Liu, Xiao and Cheng, Shuiqi and Gao, Jiyang and Xu, Huazhe and Zhao, Hang},
  journal={arXiv preprint arXiv:2509.00576},
  year={2025}
}

@article{gr3,
  title={Gr-3 technical report},
  author={Cheang, Chilam and Chen, Sijin and Cui, Zhongren and Hu, Yingdong and Huang, Liqun and Kong, Tao and Li, Hang and Li, Yifeng and Liu, Yuxiao and Ma, Xiao and others},
  journal={arXiv preprint arXiv:2507.15493},
  year={2025}
}

\end{document}